\documentclass[conference,letterpaper]{IEEEtran}
\IEEEoverridecommandlockouts
\usepackage{amsmath,amsfonts,amssymb}
\usepackage{algorithmic}
\usepackage{array}
\usepackage[caption=false,font=normalsize,labelfont=sf,textfont=sf]{subfig}
\usepackage{textcomp}
\usepackage{stfloats}
\usepackage{url}
\usepackage{verbatim}
\usepackage{graphicx}
\usepackage{cite}
\usepackage{booktabs}
\usepackage{balance}
\def\BibTeX{{\rm B\kern-.05em{\sc i\kern-.025em b}\kern-.08em
    T\kern-.1667em\lower.7ex\hbox{E}\kern-.125emX}}

\begin{document}
\topmargin=-15mm

\title{Robust Fleet Sizing for Multi-UAV Inspection Missions under Synchronized Replacement Demand}

\author{
\IEEEauthorblockN{Vishal Ramesh and Antony Thomas}
\IEEEauthorblockA{Robotics Research Center, IIIT Hyderabad, Hyderabad 500032, India}
}

\maketitle

\begin{abstract}
Multi-UAV inspection missions require spare drones to replace active drones during recharging cycles. Existing fleet-sizing approaches often assume steady-state operating conditions that do not apply to finite-horizon missions, or they treat replacement requests as statistically independent events. The latter provides per-request blocking guarantees that fail to translate to mission-level reliability when demands cluster. This paper identifies a structural failure mode where efficient routing assigns similar workloads to each UAV, leading to synchronized battery depletion and replacement bursts that exhaust the spare pool even when average capacity is sufficient. We derive a closed-form sufficient fleet-sizing rule, $k = m(\lceil R \rceil + 1)$, where $m$ is the number of active UAVs and $R$ is the recovery-to-active time ratio. This additive buffer of $m$ spares absorbs worst-case synchronized demand at recovery-cycle boundaries and ensures mission-level reliability even when all UAVs deplete simultaneously. Monte Carlo validation across five scenarios ($m \in [2, 10]$, $R \in [0.87, 3.39]$, 1000 trials each) shows that Erlang-B sizing with a per-request blocking target $\varepsilon = 0.01$ drops to 69.9\% mission success at $R = 3.39$, with 95\% of spare exhaustion events concentrated in the top-decile 5-minute demand windows. In contrast, the proposed rule maintains 99.8\% success (Wilson 95\% lower bound 99.3\%) across all tests, including wind variability up to $\mathrm{CV} = 0.30$, while requiring only four additional drones in the most demanding scenario.
\end{abstract}


\section{Introduction}
\IEEEPARstart{B}{attery-limited} endurance poses a fundamental bottleneck in multi-UAV inspection missions. In applications such as infrastructure inspection, precision agriculture, and environmental monitoring, a team of UAVs must service a finite set of distributed sites whose aggregate workload exceeds the endurance of any single vehicle~\cite{5,6}. Sustained operation therefore necessitates periodic withdrawal of active UAVs for battery replenishment, with immediate replacement by spare units drawn from a shared fleet. If a replacement vehicle is unavailable at the time of request, mission continuity is compromised. Consequently, determining the minimum number of spare UAVs required to guarantee uninterrupted task execution constitutes a critical planning and resource allocation problem.

Existing fleet-sizing approaches typically rely on one of two modeling assumptions. The first is steady-state operation, where vehicles cycle indefinitely and spare capacity follows from average utilization~\cite{1,2,3}. The second treats replacement requests as statistically independent, sizing the spare pool so each request is served with probability at least $1-\varepsilon$~\cite{4}. Both assumptions are ill-suited to finite-horizon inspection missions. Steady-state models do not account for transient demand in missions where each task is completed exactly once. Independence-based models guarantee per-request blocking but not mission-level reliability. Even $\varepsilon$ = 0.01 yields mission success well below 95\% over dozens of handovers.

A key observation of this work is that finite inspection missions induce synchronized, rather than independent, replacement demand. Energy-aware routing assigns UAVs to spatially clustered sites, leading to similar travel distances and comparable battery depletion rates across vehicles. Consequently, replacement requests occur in temporal bursts instead of being uniformly distributed over time. These synchronized bursts can temporarily exhaust the spare pool even when average utilization suggests adequate capacity. In this setting, mission failure depends on whether all active vehicles deplete simultaneously rather than on how many replacements are needed on average. The interplay between endurance and turnaround is captured by the recovery ratio $R = \bar{T}_{\mathrm{recovery}} / \bar{T}_{\mathrm{active}}$, where $\bar{T}_{\mathrm{recovery}}$ aggregates the return flight, battery recharge, and preparation for relaunch, while $\bar{T}_{\mathrm{active}}$ is the usable flight time per sortie. A larger $R$ keeps each UAV out of service longer after every sortie and therefore demands more spares to cover its absence. Fig.~1 illustrates this operational cycle and the resulting fleet-sizing challenges.

\begin{figure}[t]
  \centering
  \includegraphics[width=0.98\linewidth]{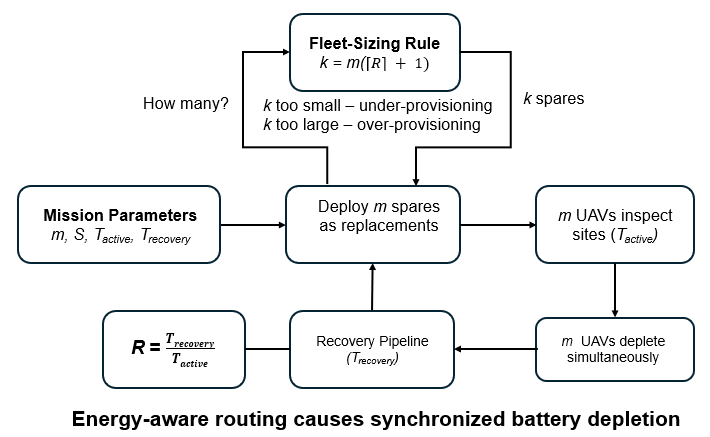}
  \caption{Operational cycle of a multi-UAV inspection mission. Energy-aware routing assigns similar workloads, producing synchronized battery depletion. The fleet-sizing rule determines the spare count $k$ from mission parameters $m$ and $R = \bar{T}_{\mathrm{recovery}} / \bar{T}_{\mathrm{active}}$.}
  \label{fig:intro_overview}
\end{figure}

This paper makes three main contributions:
\begin{enumerate}
    \item We identify synchronized replacement bursts as the fundamental structural failure mode in finite-horizon inspection missions and derive a worst-case bound on the number of concurrent recoveries under adversarial phase alignment.
    \item We propose a closed-form sufficient fleet-sizing rule, $k = m(\lceil R \rceil + 1)$, in which an additive buffer of $m$ spares absorbs worst-case demand at recovery-cycle boundaries. The rule is analytically derived and requires neither simulation, distributional assumptions, nor iterative computation.
    \item We validate the proposed sizing rule through Monte Carlo simulation across five scenarios ($m \in [2,10]$, $R \in [0.87,3.39]$, 1000 trials per setting). While Erlang-B sizing degrades to 69.9\% mission success at $R = 3.39$, the proposed rule maintains 99.8\% success (Wilson 95\% lower bound 99.3\%) with only four additional UAVs in the most demanding scenario.
\end{enumerate}
\section{Related Work}
Existing approaches to UAV fleet sizing and replacement can be organized along three dimensions: how the operational horizon is modeled, what statistical assumptions govern replacement demand, and whether fleet capacity is treated as a decision variable or a given input. We review each in turn and identify the gap that motivates the present work.

\subsection{Persistent Monitoring and Scheduling}
A substantial body of work studies how to maintain continuous UAV service over long or indefinite horizons. Song et al. \cite{1} propose a Mixed Integer Linear Program formulation for persistent UAV scheduling with split jobs, coordinating multiple UAVs and service stations to decide when each vehicle should service, transfer responsibility, and refuel. Maini et al. \cite{2} extend this setting to heterogeneous UAV--UGV teams operating under visibility constraints, jointly optimizing watchman routes and refueling stops through a branch-and-cut framework. Hartuv et al. \cite{3} consider the minimum spare drone problem more directly, casting it as a variant of the bin-packing problem and developing a greedy approximation algorithm that guarantees no battery drains during persistent monitoring with a 1.5 approximation ratio in the offline case. Battery assignment and scheduling for drone delivery is addressed in~\cite{7}, optimizing charger allocation under cyclic recovery constraints tied to energy turnaround time. More recently, Asghar et al.~\cite{21} minimize the number of robots required to satisfy latency constraints in persistent monitoring while guaranteeing periodic recharging at a depot. The formulation nonetheless remains tied to cyclic, infinite-horizon operation with per-robot recharging, rather than a shared spare pool serving synchronized demand in a finite-horizon mission.

All of the above formulations rely on a cyclic steady-state assumption, wherein UAVs cycle indefinitely and replacement demand settles into a repeating pattern. This is a structural assumption, since fleet size is derived from workload balance over the repeating tours and optimality depends on that periodic structure. In contrast, finite-horizon inspection missions differ fundamentally. Each site is visited once and the mission ends. The workload is front-loaded, so replacement demand depends on the chosen routes and does not converge to a periodic pattern. Consequently, demand exhibits non-stationary, transient bursts that cannot be represented using steady-state or cyclic workload models. This work addresses fleet sizing in such finite-horizon settings, where replacement demand is transient, mission-dependent, and governed by the one-time execution of the inspection task set.

\subsection{Statistical Resource Sizing}
An alternative perspective models the spare pool as a shared resource and applies classical queuing loss models. The Erlang-B formula, originally developed for telephone circuit provisioning \cite{4} and studied extensively by Jagerman~\cite{8}, gives the blocking probability in a system with $k$ servers and Poisson arrivals at offered load $a = \lambda s$, where $\lambda$ is the arrival rate and $s$ is the mean service time. The model belongs to the broader class of loss networks analyzed by Kelly~\cite{9}, in which blocked requests are cleared without retry. In the UAV setting, the offered load, defined as the average number of UAVs needing recovery at any time, is $a = mR$, with $m$ active UAVs and $R$ the recovery-to-active time ratio (see Section~\ref{sec:experiments}). The spare pool size is chosen as the smallest $k$ such that the blocking probability $B(k, mR)$ does not exceed a target $\varepsilon$. A blocked replacement request corresponds to the absence of an available spare UAV when a replacement is needed. The Erlang-B formula computes the steady-state probability of such blocking given $k$ available spares and offered load $a = mR$. Standard engineering practice sizes $k$ so that this per-request blocking probability does not exceed a small threshold. We adopt the conventional target $\varepsilon = 0.01$, consistent with loss-network provisioning practice in telecommunications \cite{8, 9}, yielding a 99\% per-request service guarantee.

Two structural issues arise for finite inspection missions. First, Erlang-B provides a per-request guarantee. It bounds the probability that a single replacement request is blocked. Mission success, however, requires that every handover succeed. Even under independence, the probability of zero blocked events over $H$ handovers scales as $(1 - \varepsilon)^H$, deteriorating rapidly with mission length. Per-request sizing therefore does not translate into mission-level reliability. Second, the independence assumption itself is poorly matched to this domain. Loss models assume each replacement request arrives independently of all others, whereas battery discharge is governed by known flight profiles and route lengths. Later work allows arrival rates to change over time~\cite{10} and relaxes the independence assumption~\cite{11}, but still treat replacement requests as random. In our setting, replacements are not random and are fixed by the routes and battery limits. Bandi et al.~\cite{12} develop a robust queueing framework that replaces distributional assumptions with worst-case bounds on arrival and service processes. Their approach motivates the shift from statistical blocking guarantees to deterministic capacity provisioning, though they do not address the spatial synchronization structure specific to UAV fleet operations. However, none of these formulations provide mission-level sufficiency guarantees that account for the correlated replacement demand induced by energy-aware routing in finite inspection missions.
\subsection{UAV Fleet Coordination and Replacement}
Another line of work focuses on the mechanics of UAV replacement rather than on spare pool sizing. Erdelj et al. \cite{13} develop a reactive continuity-of-service algorithm implemented via MAVLink extensions, in which a depleting UAV broadcasts a request and the ground control station assigns the nearest available spare. The method however assumes availability of sufficient spares. Broader work on collaborative air-ground systems \cite{14} and risk-aware UAV-UGV rendezvous planning \cite{15} studies coordination and refueling logistics for heterogeneous teams, yet treats fleet capacity as given rather than as a decision variable.

In summary, scheduling approaches rely on cyclic steady state, queueing models assume stochastic independence, and coordination algorithms assume fixed capacity. None directly addresses spare fleet sizing for finite-horizon missions with synchronized replacement demand. Where prior work determines when to replace, how to replace, or how likely a single request is blocked, we address how many spares are sufficient for mission-level success. We consider predictable battery discharge driven by known flight profiles, energy-aware routing that causes replacement requests to cluster in time, and a mission-level reliability requirement. Our objective is to determine the minimum spare fleet size that ensures zero spare exhaustion events over the entire mission horizon.
\subsection{The Fleet-Sizing Gap}
Across the three preceding categories, a consistent pattern emerges. Scheduling methods \cite{1,2,3,7,21} derive fleet requirements from cyclic workload balance under indefinite operation. Statistical models \cite{4,8,9,10,11,12} provide per-request blocking guarantees under independence assumptions. Coordination and replacement algorithms \cite{13,14,15} assume that sufficient spares are available and focus on the mechanics of handover execution rather than on how many spares are needed. No existing method addresses fleet sizing for finite-horizon missions in which replacement demand is neither periodic nor stochastic but deterministic and geometry-driven. In such missions, energy-aware routing assigns spatially clustered workloads of similar size, producing battery depletion schedules that are largely predictable from route structure. The resulting synchronized replacement waves represent a fundamentally different type of demand compared to the independent, random arrivals assumed by prior work.

This paper addresses this gap by deriving a fleet-sizing rule based on worst-case synchronization analysis, rather than steady-state or probabilistic assumptions, and by validating sufficiency with respect to mission-level success criteria rather than per-request blocking probabilities.
\section{Problem Formulation}
\label{sec:PF}
\subsection{System Model}
We consider a fleet of homogeneous UAVs tasked with inspecting a bounded geographic region $A$, partitioned into a finite set of discrete inspection sites $S$. The fleet comprises $m$ active UAVs performing inspection tasks and $k$ spare UAVs maintained in a shared pool at a central base. 

Each active UAV follows a pre-computed, energy-aware route visiting a subset of clustered sites, performing a sensing task of fixed duration $t_{\text{scan}}$ at each location.

All vehicles share identical battery capacity and flight characteristics. The usable battery capacity defines a uniform maximum active time $T_{\text{active}}$. Once this limit is reached, the UAV must return to base. Homogeneity is the worst case for synchronization, since identical capacities and launch times align depletions exactly. Heterogeneous platforms desynchronize naturally, which can only relax the buffer requirement rather than violate it. The recovery process, which includes return flight, landing, battery recharge, and preparation for relaunch, requires a total time $T_{\text{recovery}}$. We define the recovery ratio as
\begin{equation}
   R = T_{\text{recovery}} / T_{\text{active}} 
\end{equation}
When an active UAV reaches a withdrawal threshold, it requests a replacement from the spare pool. If no spare is available, a spare exhaustion event occurs. Charging capacity is assumed unlimited, and the only constraint is the number of flight-ready spares available at the moment a request arrives. This is a deliberate isolation choice. By holding charger capacity non-binding, the spare inventory becomes the sole capacity constraint, so that failures are attributable to synchronized demand rather than confounded with charger contention.

The mission objective is to inspect every site in $
S$ exactly once. Since the total workload exceeds the endurance of any single UAV, the mission requires $m$ concurrently active vehicles with periodic replacement from the spare pool. We now formally define the problem addressed in this work:

\textit{Given the number of active UAVs $m$ and the recovery ratio R, determine  the minimum spare fleet size $k$ required to ensure that no replacement request goes unserved over the entire mission horizon.}
\subsection{Mission Success Criterion}
 As the site locations within region $A$ and travel times are modeled as random variables, mission execution is stochastic. For a fixed fleet configuration $(m, k)$, different realizations of site distributions and travel-time variability may lead to different operational outcomes. We therefore evaluate performance using Monte Carlo simulation. A \textit{trial} is defined as a single complete mission execution under a fixed fleet configuration $(m, k)$ and one random realization of site locations and travel times. A trial is declared successful if: (1) All sites in $S$ are inspected exactly once, and (2) No spare exhaustion event occurs at any time during the mission. A spare exhaustion event, equivalent to a blocked request in the queueing terminology of Section~II-B, denotes a single replacement request that finds no flight-ready spare. We distinguish this single-request event from mission failure, which is the whole-mission outcome. Because inspection must proceed without interruption, a single spare exhaustion event causes mission failure, and per-request blocking guarantees therefore do not imply mission-level reliability.

Mission reliability is estimated over $N = 1000$ independent trials. We report both the empirical success rate and the Wilson 95\% lower confidence bound~\cite{16}. The Wilson interval is preferred over the Wald interval for its better coverage of bounded proportions near extreme values~\cite{17}. A fleet configuration is certified as reliable if the lower confidence bound is at least 0.95. This ensures the true mission success probability exceeds 95\% with 95\% confidence, a statistically rigorous criterion for fleet sizing.
\subsection{Baseline Sizing Methods}
We compare four fleet-sizing rules that represent progressively refined assumptions about replacement demand.

\begin{itemize}
    \item \textbf{Naive.} $k=m$. Each active UAV has one dedicated spare. Recovery time is ignored, so for $R > 1$ the spare may still be charging when the next replacement is required.
    \item \textbf{Duty-Cycle.} $k = m \cdot \lceil R \rceil$. Each active position is supported by $\lceil R \rceil$ spares, which is sufficient to maintain continuous rotation through the recovery pipeline under perfectly staggered timing. This accounts for recovery duration but does not provision for simultaneous demand from multiple active positions.
    \item \textbf{Erlang-B.} $k = \min \{ k \mid B(k, mR) \le \varepsilon \}$, with $\varepsilon = 0.01$. The offered load is $a = mR$, and the spare pool is sized so that the per-request blocking probability does not exceed 1\%. This is the standard application of the Erlang-B loss model under an independence assumption.
    \item \textbf{Proposed.} $k = m \cdot (\lceil R \rceil + 1)$. This adds a buffer of $m$ spares beyond the Duty-Cycle allocation. The derivation and justification of this rule are presented in Section IV.
\end{itemize}


\section{Sufficient Buffered Fleet-Sizing Rule}

\subsection{Worst-Case Concurrent Recoveries}

We begin by bounding the maximum number of UAVs that can be simultaneously in recovery at any instant during the mission. We use the term \textit{phase alignment} to describe how closely the battery depletion times of different UAVs coincide. Worst-case phase alignment is the scenario where all $m$ active UAVs deplete at the same instant, producing simultaneous replacement requests.

\textbf{Proposition~1:} \textit{Under worst-case phase alignment, up to $m\lceil R \rceil$ UAVs may be simultaneously recovering.}

\textit{Proof.}
Let $C(t)$ denote the number of UAVs in recovery at time~$t$.
All $m$ active UAVs launch at $t=0$ and deplete at $t=T_{\mathrm{active}}$,
entering recovery simultaneously, so $C(T_{\mathrm{active}})=m$.
We refer to each group of $m$ simultaneous replacement requests as a \textit{replacement wave}. At $t = j\,T_{\mathrm{active}}$ for $j=1,2,\ldots,\lceil R\rceil$,
the $j$-th replacement cohort of $m$~UAVs completes its active phase and enters recovery.
The earliest cohort, which entered recovery at $t=T_{\mathrm{active}}$,
completes recovery at $t=(1+R)\,T_{\mathrm{active}}$.
For all $j \leq \lceil R \rceil$, we have
$j\,T_{\mathrm{active}} \leq (1+R)\,T_{\mathrm{active}}$,
so no cohort has exited recovery when the $\lceil R \rceil$-th cohort enters.
Therefore $C(\lceil R \rceil T_{\mathrm{active}}) = m\lceil R \rceil$.
Between wave arrivals, no new cohort enters recovery; cohorts
only exit. At most $\lceil R \rceil$ cohorts can overlap in recovery before
the earliest exits, so $C(t) \leq m\lceil R \rceil$ for all~$t$.
\hfill$\blacksquare$
\subsection{Boundary-Burst Mechanism}
Proposition~1 establishes that the recovery pipeline can accommodate at most $m\lceil R \rceil$ UAVs. The Duty-Cycle rule provisions exactly this number of spares. While this matches the maximum concurrent recoveries, it does not guarantee feasibility.

At time $\lceil R \rceil \, T_{\mathrm{active}}$, all $m\lceil R \rceil$ spares are committed to recovery. At that same instant, the active cohort of $m$~UAVs completes its duty cycle and requests replacement. Under Duty-Cycle sizing the pool returns to exactly empty at this boundary, so the schedule is feasible in the noise-free limit, where the earliest recovering cohort returns just before the next wave arrives. The allocation therefore carries zero count margin. The system needs $m\lceil R \rceil$ UAVs to populate the recovery pipeline and $m$ flight-ready UAVs to absorb a wave arriving before the earliest cohort has recovered. The Duty-Cycle allocation provides only the former, so any perturbation that delays a recovering UAV past the next wave produces exhaustion. This vulnerability grows with $m$ and $R$, since larger synchronized waves and longer recovery leave less tolerance for timing variation. The imbalance is resolved by provisioning an additional buffer of $m$ spares. Specifically, with
\begin{equation}
k = m(\lceil R \rceil + 1),
\end{equation}
the system contains $m$ UAVs beyond the maximum recovery pipeline occupancy. These additional UAVs remain flight-ready and absorb the incoming replacement wave, while the recovery pipeline continues at full capacity without interruption. The buffer therefore functions as a robustness margin rather than a correction to a deterministic deficit. For small $m$ and moderate $R$, the zero-margin Duty-Cycle schedule remains feasible, consistent with the experimental results in Section~\ref{sec:experiments}, where Duty-Cycle certifies in S1 through S3 and fails only in the high-$m$, high-$R$ scenarios S4 and S5.
\subsection{Fleet-Sizing Rule}
\textbf{Proposition~2:} \textit{For $m$ active UAVs and recovery ratio $R$, a spare fleet of size}
\vspace{-0.2cm}
\begin{equation}
k = m(\lceil R \rceil + 1)
\end{equation}
\textit{is sufficient to prevent spare exhaustion under worst-case phase alignment.}

\textit{Proof.}
Define the flight-ready spare count
$F(t) = k - C(t)$,
where $C(t)$ is the number of UAVs in recovery at time~$t$.
A replacement wave is served without exhaustion if and only if
$F(t) \geq m$ at the moment the wave arrives.

\textit{Base case.}
At $t = \lceil R \rceil \cdot T_{\mathrm{active}}$,
Proposition~1 gives $C(t) = m\lceil R \rceil$.
With $k = m(\lceil R \rceil + 1)$, the flight-ready count is
$F(t) = m(\lceil R \rceil + 1) - m\lceil R \rceil = m$.
The boundary wave of $m$ requests is served exactly.

\textit{Subsequent waves.}
At each subsequent wave time $t = j\,T_{\mathrm{active}}$ for $j > \lceil R \rceil$,
the cohort that entered recovery at
$t = (j - \lceil R \rceil)\,T_{\mathrm{active}}$
has been recovering for
$\lceil R \rceil\,T_{\mathrm{active}} \geq R\,T_{\mathrm{active}}$
and has therefore completed recovery, returning $m$~UAVs to the pool.
By Proposition~1, at most $m\lceil R \rceil$ UAVs are in recovery at any
wave time. With $k = m(\lceil R \rceil + 1)$, the flight-ready count satisfies
$F(j\,T_{\mathrm{active}}) \geq k - m\lceil R \rceil = m$
for every wave~$j$. No spare exhaustion occurs. \hfill$\blacksquare$

The worst-case phase alignment considered here is sufficient but not necessary for failure, as the rule protects against synchronized replacement waves rather than arbitrary adversarial timing. This synchronized regime is the one induced by the offline pre-planned launches and identical batteries assumed throughout. Online replanning would constitute a different operating regime that could partially de-synchronize depletion, in which case the worst-case alignment is not reached and the rule becomes conservative rather than unsafe. Our guarantee is therefore scoped to the synchronized regime, where it is most needed.

\textbf{Corollary~1} (Low-Recovery Regime): \textit{If $R < 1$, then $k = m$ is sufficient to prevent spare exhaustion under worst-case phase alignment.}

\textit{Argument.} When $R < 1$, recovery completes before the next wave arrives, so at most $m$ vehicles are simultaneously recovering.

\textbf{Corollary~2} (Perfectly Staggered Regime): \textit{If replacement requests are perfectly staggered such that each active position is offset by $T_{\mathrm{active}}/\lceil R \rceil$, then $k = m\lceil R \rceil$ is sufficient.}

\textit{Interpretation.} In practice, energy-aware routing assigns similar workloads to vehicles launched simultaneously, making perfect staggering difficult without sacrificing routing efficiency. The buffer therefore addresses the synchronization that arises naturally from efficient task allocation.

The resulting rule is closed-form and directly computable from $m$ and~$R$, requiring no simulation and no distributional assumptions. Our focus is on sufficiency rather than optimality, consistent with the robust optimization approach~\cite{18}, which plans for the worst case. Here the uncertainty is how closely the UAVs deplete simultaneously, and the rule ensures feasibility regardless of that alignment. The ceiling function introduces conservatism near integer values of~$R$. For example, as $R$ crosses an integer threshold, the required fleet increases by $m$~UAVs. Tightening this discontinuity while retaining worst-case guarantees is a direction for future work. As shown in Section~\ref{sec:experiments}, however, the resulting margin remains modest in practical scenarios. 
\subsection{Per-Request vs.\ Mission-Level Reliability}
\begin{figure}[t]
    \centering
    \includegraphics[width=0.85\linewidth]{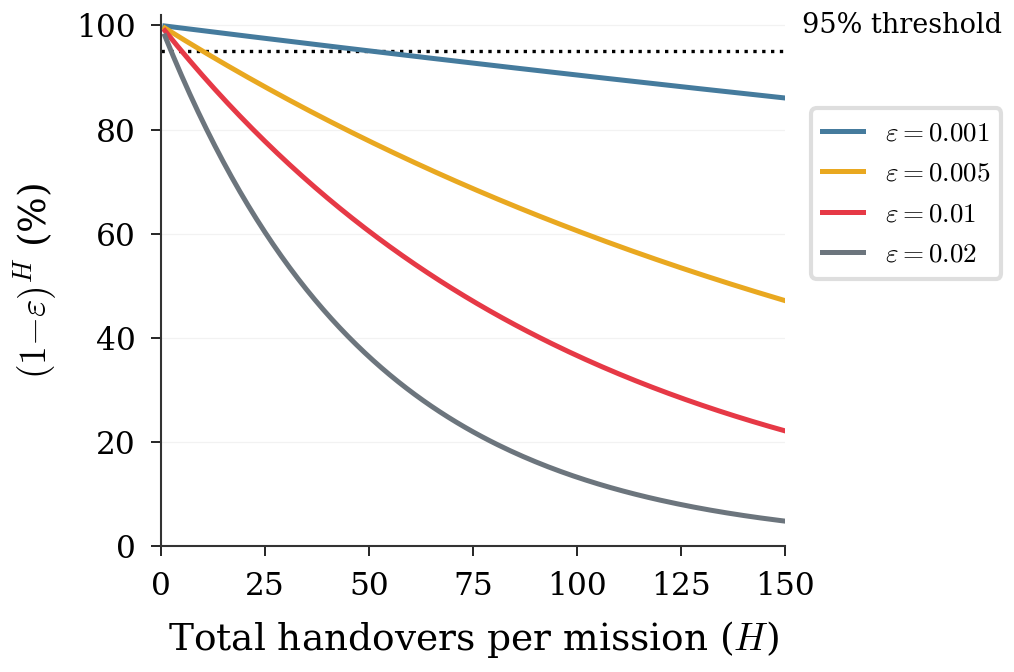} 
    \caption{Mission success probability under independent per-request blocking as a function of the number of handovers $H$. Even with a 1\% blocking target ($\varepsilon = 0.01$), mission-level reliability deteriorates rapidly as $H$ increases.}
    \label{fig:reliability_compounding}
\end{figure}

Even if independence were satisfied, per-request blocking guarantees do not ensure mission-level reliability. Under Erlang-B with blocking target~$\varepsilon$, each individual handover succeeds with probability at least $1 - \varepsilon$. For a mission requiring $H$~handovers, independence implies
\begin{equation}
P(\text{mission success}) = (1 - \varepsilon)^{H}
\end{equation}
Figure~\ref{fig:reliability_compounding} plots this relationship as a function of $H$ for $\varepsilon \in \{0.001, 0.005, 0.01, 0.02\}$. Even with $\varepsilon = 0.01$, mission success falls below 95\% at $H = 5$ and below 50\% by $H = 69$. Our experimental missions require between 6 and 52 handovers, placing them well within the regime where per-request guarantees offer little assurance at the mission level.

The proposed rule eliminates this by targeting zero spare exhaustion through worst-case capacity provisioning.
\section{Experimental Validation}
\label{sec:experiments}
\subsection{Simulation Framework}

The simulation framework accepts user-specified mission parameters: the number of inspection sites S, operating area dimensions, base station position, battery endurance, charging duration, scan time per site, and flight speed. The number of active UAVs m is determined from the ratio of total inspection workload to per-sortie capacity, and the spare count k is computed by the fleet-sizing rule under evaluation. Before execution, a feasibility check verifies that every site is reachable from the base within a single battery cycle, accounting for round-trip transit and a 15\% energy reserve; infeasible configurations are rejected.
For each trial, site positions are drawn from a clustered spatial distribution within the operating area and partitioned among active UAVs via k-means clustering. Each UAV follows a nearest-neighbor route within its assigned cluster. The simulation advances in fixed 0.5-minute timesteps. Battery depletion is tracked continuously and a replacement is requested when the remaining charge equals the estimated return-to-base time plus the safety reserve. Travel times are subject to a per-trial common-mode wind factor drawn from a log-normal distribution with unit mean and specified coefficient of variation, combined with per-leg independent uniform perturbation. Charging capacity is unlimited, so the spare inventory is the sole capacity constraint. A replacement drone inherits the departing vehicle's position, remaining task sequence, and partial scan progress. Trial seeds are shared across all sizing methods, ensuring identical site layouts and travel-time realizations so that only the spare count differs between methods.

\subsection{Experimental Setup}
We evaluate the four fleet-sizing rules using Monte Carlo simulation across five scenarios of increasing operational difficulty. Each scenario specifies the number of inspection sites, battery life, scan time per site, charge time, map dimensions, and flight speed. The number of active UAVs $m$ is determined by the ratio of total inspection workload to the per-vehicle capacity per sortie, where capacity accounts for usable battery life, scan duration, and inter-site transit. These parameters jointly determine $m$, the recovery ratio $R$, and the spare count $k$ under each sizing rule. Table~\ref{tab:results} summarizes the resulting $(m,R)$ pairs, fleet sizes, and performance outcomes.

The baseline coefficient of variation of the common-mode wind factor is set to $\mathrm{CV}=0.15$, the midpoint of the range swept in the robustness analysis of Section~V-G ($\mathrm{CV}$ from 0.00 to 0.30). Synchronization arises naturally from the shared launch time and similar route geometries.

To measure demand concentration, we partition each mission timeline into non-overlapping 5-minute windows, rank them by the number of replacement requests, and define the top-decile windows as those in the highest 10\% by request count. Each scenario is evaluated over $N{=}1000$ independent trials per method. Mission success requires that all sites are inspected and no spare exhaustion event occurs. We report empirical success rates, Wilson 95\% lower confidence bounds (WLB), and the burst-failure concentration metric defined below.

\begin{table*}[t]
\caption{Derived scenario values, fleet sizes under each sizing rule, mission success rates with Wilson 95\% lower confidence bounds, and burst-failure concentration. The sizing rules are defined as follows: Naive ($k = m$), Duty-Cycle ($k = m\lceil R\rceil$), Erlang-B ($k = \min{k \mid B(k, mR) \le 0.01}$), and Proposed ($k = m(\lceil R\rceil + 1)$). $\bar{H}$ denotes the mean number of handovers per successful trial under the proposed method. Burst-failure concentration (\%) reports the fraction of spare-exhaustion events falling within top decile 5-minute demand windows. `--' indicates that no exhaustion events were observed.}
\label{tab:results}
\centering
\scriptsize
\setlength{\tabcolsep}{2.6pt}
\renewcommand{\arraystretch}{1.0}
\scalebox{0.97}{
\begin{tabular*}{\textwidth}{@{\extracolsep{\fill}} l ccc cccc cccc cccc cc}
\toprule
& & & 
& \multicolumn{4}{c}{Fleet Size ($k$)} 
& \multicolumn{4}{c}{Mission Success Rate} 
& \multicolumn{4}{c}{Wilson 95\% Lower Confidence Bound} 
& \multicolumn{2}{c}{Burst-Failure Concentration (\%)} \\
\cmidrule(lr){5-8} \cmidrule(lr){9-12} \cmidrule(lr){13-16} \cmidrule(lr){17-18}
 & $m$ & $R$ & $\bar{H}$ 
 & Naive & Duty-Cycle & Erlang-B & Proposed
 & Naive & Duty-Cycle & Erlang-B & Proposed
 & Naive & Duty-Cycle & Erlang-B & Proposed
 & Duty-Cycle & Erlang-B \\
\midrule
S1 & 2  & 0.87 &  5.6 &  2 &  2 &  6 &  4 & 1.000 & 1.000 & 1.000 & 1.000 & 0.996 & 0.996 & 0.996 & 0.996 & ---  & ---   \\
S2 & 2  & 1.59 &  7.0 &  2 &  4 &  9 &  6 & 0.000 & 1.000 & 1.000 & 1.000 & 0.000 & 0.996 & 0.996 & 0.996 & ---  & ---   \\
S3 & 4  & 2.15 & 24.6 &  4 & 12 & 16 & 16 & 0.000 & 1.000 & 1.000 & 1.000 & 0.000 & 0.996 & 0.996 & 0.996 & ---  & ---   \\
S4 & 7  & 3.30 & 46.9 &  7 & 28 & 34 & 35 & 0.000 & 0.136 & 0.997 & 1.000 & 0.000 & 0.116 & 0.991 & 0.996 & 94.5 & 100.0 \\
S5 & 10 & 3.39 & 52.3 & 10 & 40 & 46 & 50 & 0.000 & 0.002 & 0.699 & 0.998 & 0.000 & 0.001 & 0.670 & 0.993 & 82.2 & 95.0  \\
\bottomrule
\end{tabular*}}
\end{table*}
\subsection{Mission Reliability}
\begin{figure}[t]
    \centering
    \includegraphics[width=\linewidth]{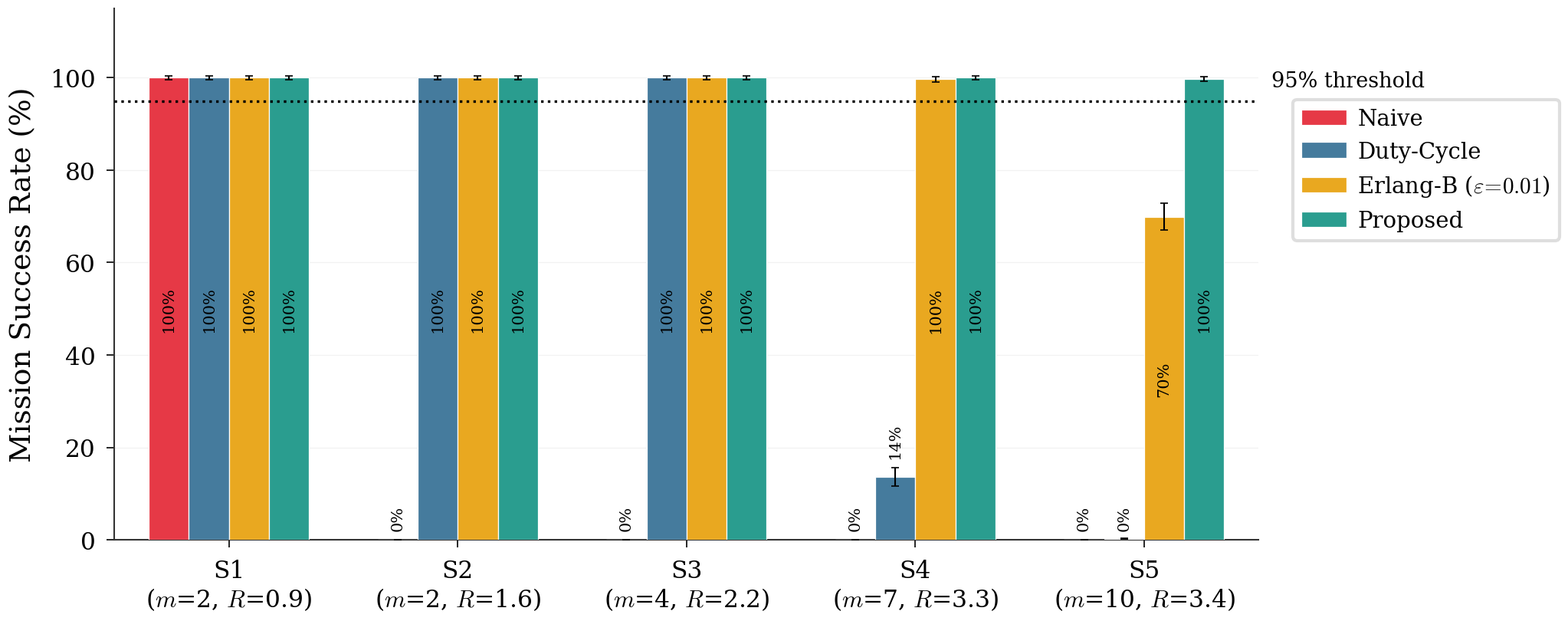}
    \caption{Mission success rates with Wilson 95\% confidence intervals across five scenarios. The 95\% reliability certification threshold is indicated for reference.}
    \label{fig:mission_success}
\end{figure}

Fig.~\ref{fig:mission_success} shows mission success rates with Wilson 95\% confidence intervals across all five scenarios. In $S1$ ($m=2, R=0.87$), all four methods pass certification. 
When $R < 1$, even the Naive allocation $k=m$ provides enough spares to cover all replacement requests.

In $S2$ ($m=2, R=1.59$) and $S_3$ ($m=4, R=2.15$), Naive allocation fails  certification while Duty-Cycle, Erlang-B, and Proposed continue to pass. These scenarios mark the onset of sustained replacement pressure ($R > 1$), where minimal redundancy is no longer sufficient. However, at these fleet scales, Duty-Cycle and Erlang-B still maintain adequate feasibility margins for certification-level reliability.

In $S4$ ($m=7, R=3.30$), Duty-Cycle fails certification (WLB 11.6\%), with empirical success falling to 13.6\%. The ceiling function yields $\lceil 3.30 \rceil = 4$, resulting in an allocation of $k=28$ spares. This exactly matches the maximum recovery pipeline capacity predicted by Proposition 1, placing Duty-Cycle at the zero-margin boundary identified in Section IV-B, where any timing perturbation tips the schedule into exhaustion. Erlang-B and Proposed both pass certification.

In $S5$ ($m=10, R=3.39$), Erlang-B fails certification (WLB 67.0\%), achieving only 69.9\% empirical success. Proposed passes certification (WLB 99.3\%), achieving 99.8\% empirical success. Erlang-B allocates $k=46$ spares, four fewer than the Proposed fleet of $k=50$. This difference of four produces a qualitative reliability transition: Erlang-B operates below the feasibility threshold, while Proposed remains safely above it. Duty-Cycle fails almost completely, achieving only 0.2\% success (WLB 0.1\%).

Overall, mission reliability is governed by feasibility thresholds rather than gradual degradation. Baseline methods perform adequately only within system capacity limits, and once these limits are reached, reliability collapses abruptly.

\subsection{Per-Request vs.\ Mission-Level Gap}
\begin{figure}[t]
    \centering
    \includegraphics[width=0.85\linewidth]{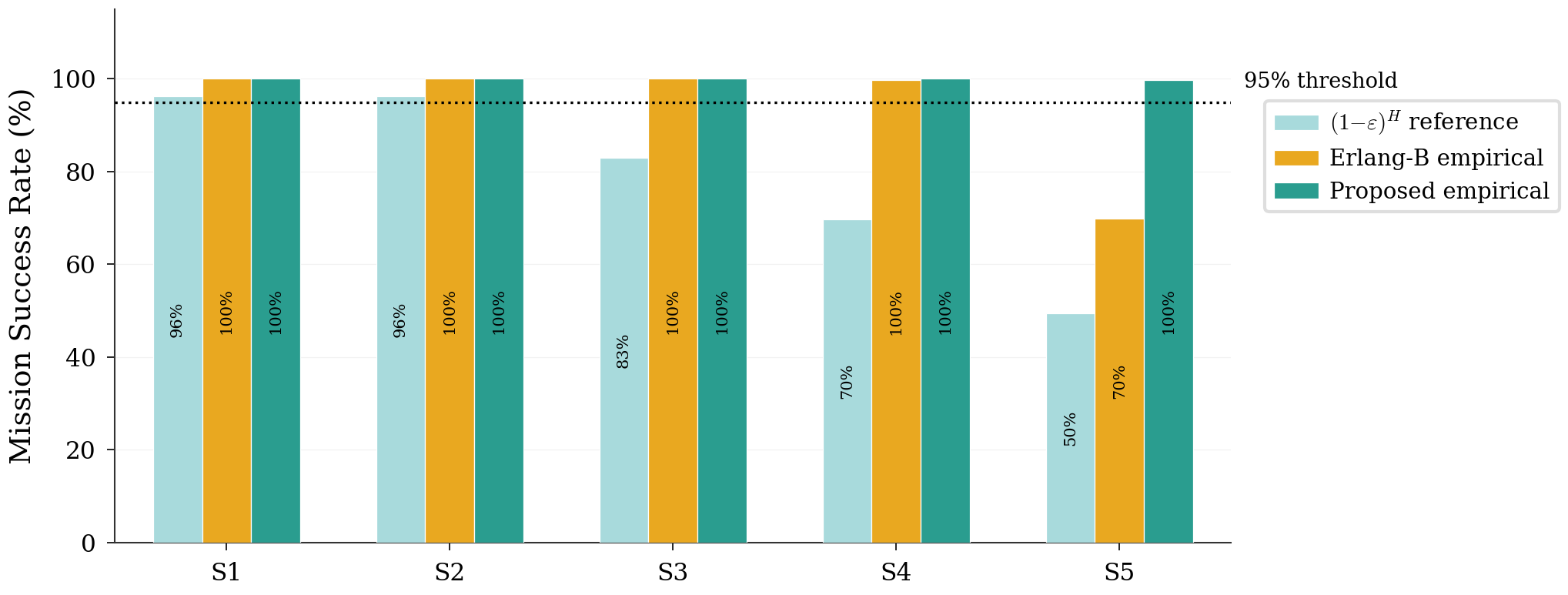}
    \caption{Erlang-B empirical mission success compared with the independence reference $(1-\varepsilon)^{H}$, computed from the per-request blocking target and observed handover count.}
    \label{fig:erlang_vs_reference}
\end{figure}

Fig.~\ref{fig:erlang_vs_reference} compares Erlang-B's empirical mission success against the independence reference $(1-\varepsilon)^{H}$, using the mean handover count $H$ from Table~I, obtained under the Proposed method, as the common mission length for all methods.

In S1 through S3, both the empirical rate and the reference exceed 95\%, so per-request blocking is not yet limiting mission reliability. In S4, the independence reference drops below 95\% as $H$ increases, consistent with the compounding effect derived in Section IV-D. In S5, the reference falls to approximately 44\%, while Erlang-B achieves 69.9\%. The empirical rate exceeds the reference in every scenario since blocked UAVs leave the system and reduce subsequent replacement demand. Even so, Erlang-B does not meet the 95\% reliability threshold in the most demanding scenario.

\subsection{Synchronized Demand Structure}
To establish a direct link between synchronization and mission failure, we first examine when spare exhaustion events occur. For each trial in which exhaustion happens, we determine whether the event falls within a top-decile 5-minute demand window. In S5, 95.0\% of Erlang-B failures occur within such high-demand windows. In S4, the concentration reaches 100.0\% for Erlang-B and 94.5\% for Duty-Cycle. Failures do not accumulate gradually but concentrate during synchronized bursts of replacement requests.

\begin{figure}[t]
    \centering
    \includegraphics[width=\linewidth]{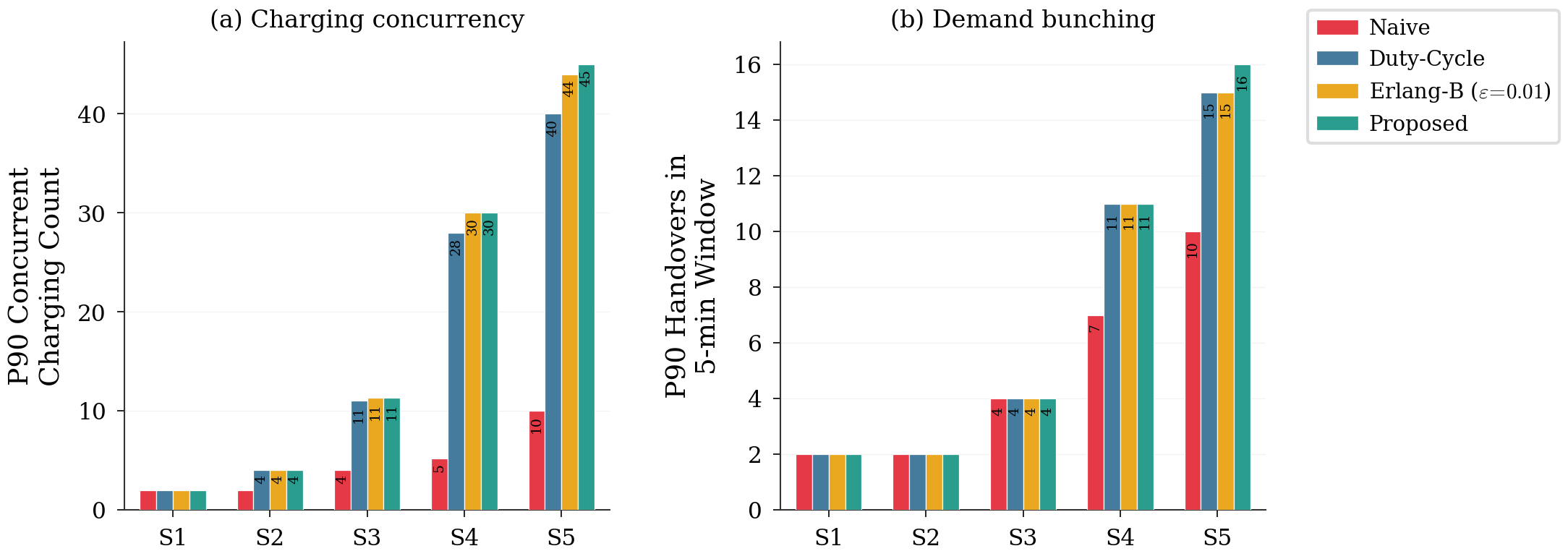}
    \caption{Evidence of synchronized replacement demand. Left: 90th-percentile number of UAVs simultaneously in recovery. Right: 90th-percentile number of handover requests within any 5-minute window.}
    \label{fig:sync_evidence}
\end{figure}

Fig.~\ref{fig:sync_evidence} provides supporting evidence for this structure. The left panel shows the 90th-percentile number of UAVs simultaneously in recovery, while the right panel shows the 90th-percentile number of handover requests within any 5-minute window. Both metrics increase with scenario difficulty. In S5, the 90th-percentile concurrent recovery count reaches approximately~45, close to $m\lceil R\rceil{=}40$, with additional overlap from travel-time variability. Demand bunching reaches 16 requests within a single 5-minute window, confirming that arrivals occur in concentrated waves. These results confirm that the dominant failure mechanism is the boundary-burst condition identified in Section IV-B.

\subsection{Handover Frequency}
\begin{figure}[t]
    \centering
    \includegraphics[width=0.85\linewidth]{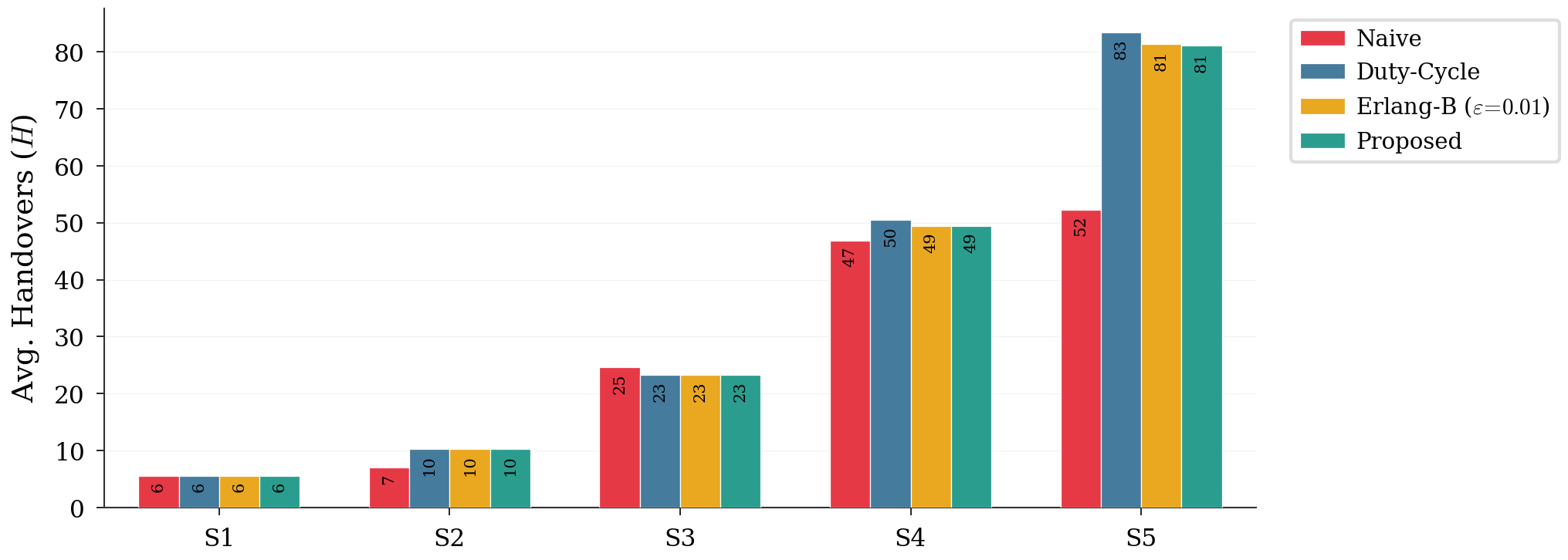}
    \caption{Average number of handovers $H$ per successful mission, by method and scenario.}
    \label{fig:handover_counts}
\end{figure}

As shown in Fig.~\ref{fig:handover_counts}, $H$ is nearly identical across Duty-Cycle, Erlang-B, and Proposed within each scenario. In S5, all three methods produce roughly $H=52$ per mission. Increasing the spare count does not reduce replacement demand, since the number of handovers is determined by $m$ and $R$, rather than the spare allocation $k$. Additional spares ensure requests are served but do not change how many are generated. This invariance justifies the use of a common $H$ when computing the independence reference in Section~V-D.

\subsection{Robustness to Wind Variability}
\begin{figure}[t]
    \centering
    \includegraphics[width=0.85\linewidth]{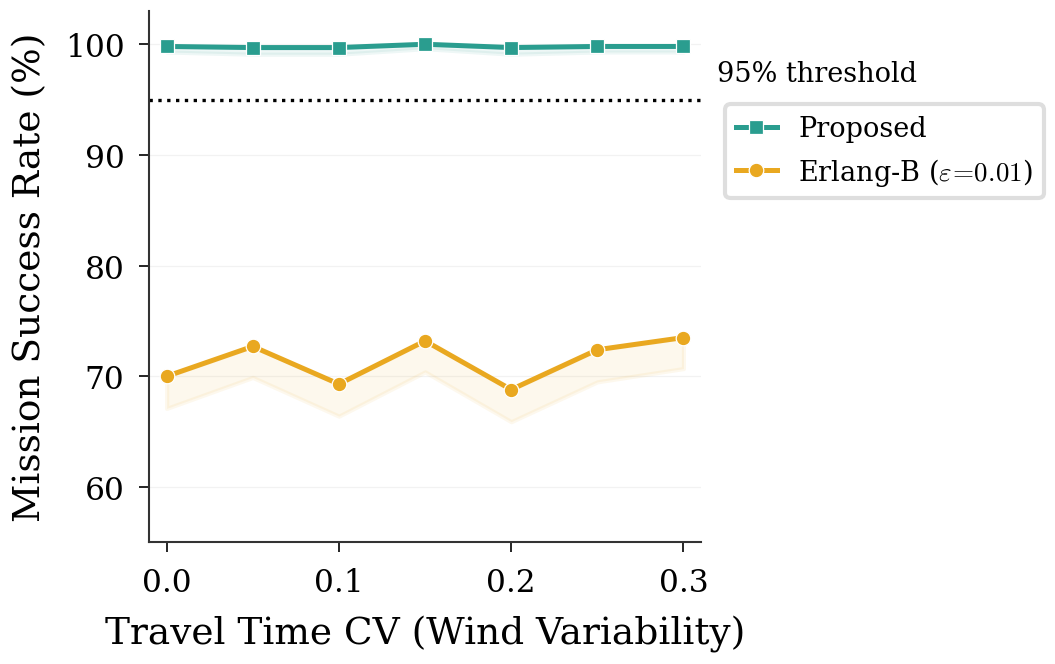}
    \caption{Mission success rates versus travel-time variability (coefficient of variation) for S5, comparing Erlang-B and Proposed fleet-sizing rules.}
    \label{fig:robustness_cv}
\end{figure}

Fig.~\ref{fig:robustness_cv} shows mission success rates for Erlang-B and Proposed across a sweep of travel-time variability from $\mathrm{CV}{=}0.00$ to~$0.30$, evaluated on S5 with 1000 trials per value. Proposed maintains a Wilson lower bound above 99\% across the full range. Erlang-B remains near 70\% regardless of variability level. The proposed rule is therefore insensitive to the magnitude of travel-time noise within realistic bounds, aligning with the theoretical development based on worst-case phase alignment rather than a specific variance model.

\section{Discussion}

\subsection{Practical Implications}
\looseness=-1
The proposed fleet-sizing rule is conservative by design. In the most demanding scenario S5, it requires 50 spares against 46 under Erlang-B, an increase of four UAVs or roughly 7\% of the fleet, raising mission reliability from 69.9\% to 99.8\%. Bertsimas and Sim~\cite{19} characterize this tradeoff as the price of robustness. Here four UAVs shift performance from failing one mission in three to fewer than one in five hundred, a modest cost for a 30 percentage point gain.

The values of $m$ and $R$ are known at planning time from battery specifications, charging duration, and workload analysis, so the spare count is computed in closed form in constant time, making it practical for field deployment.

\subsection{Scope and Limitations}
Several assumptions bound the scope of these results. We do not claim minimality of the fleet size under partial synchronization. Characterizing tight necessary bounds remains an open problem. First, active time $T_{\text{active}}$ is treated as uniform across vehicles. The simulation captures moderate variability through travel-time noise up to CV = 0.30, but strong systematic asymmetries in cluster size or dispersion could exceed the tested range. Second, charging capacity is held unlimited to isolate the spare inventory as the sole capacity constraint. Relaxing this is a natural extension. With limited charging pads, contention increases effective recovery time and therefore $R$, and because wait time depends on fleet size, this couples $R$ to the fleet and warrants separate treatment.

Third, routes are fixed prior to launch. Adaptive rerouting or dynamic workload re-balancing could partially de-synchronize replacement demand. This would reduce synchronization and likely make the rule conservative rather than unsafe, since the derivation assumes worst-case phase alignment. Fourth, the fleet is homogeneous, which is the worst case for synchronization rather than a convenience. Heterogeneous platforms yield differing recovery ratios that desynchronize depletion, so the rule remains safe and becomes conservative. A generalized treatment of non-uniform $R$ that recovers the tighter bound is left to future work.
\subsection{Future Work}
Several directions follow from this study. Incorporating stochastic active durations could reduce ceiling-function conservatism. Jointly optimizing routing and fleet size may exploit the interaction between task allocation and synchronization to achieve equivalent reliability with fewer spares. Extending the framework to heterogeneous fleets and limited charging infrastructure would broaden applicability. Formulating the problem as a robust Markov decision process \cite{20} could enable adaptive policies that adjust spare allocation while retaining worst-case guarantees.

The convergence of Erlang-B and Proposed fleet sizes in S3 suggests the existence of regime boundaries in $(m, R)$ space where simpler methods suffice and characterizing them analytically would clarify when conservative buffering is necessary.
\balance
\section{Conclusion}
This paper addressed the problem of sizing a spare UAV fleet for finite-horizon inspection missions with synchronized replacement demand. We derived a closed-form sufficient fleet-sizing rule, $
k = m(\lceil R \rceil + 1)$, based on a worst-case analysis of concurrent recovery occupancy and boundary-burst demand. Monte Carlo evaluation across five scenarios with recovery ratios ranging from 0.87 to 3.39 confirmed that the rule passes the 95\% Wilson certification threshold in all scenarios, with lower bounds exceeding 99\% in the most demanding case, while Erlang-B sizing falls to 69.9\%, requiring only four additional UAVs to close that reliability gap.

The proposed rule requires no simulation or distributional modeling and evaluates directly from system specifications, making it suitable for operational planning in battery-constrained multi-UAV settings.

\section*{ACKNOWLEDGMENT}
The authors acknowledge the support provided by MeitY,
Government of India, under the project ``SwaYaan - Capacity Building for Human Resource Development in Unmanned Aircraft System (Drone and Related Technology).''
\balance


\begin{thebibliography}{21}

\bibitem{1}
B.~D. Song, J.~Kim, J.~Kim, H.~Park, J.~R. Morrison, and D.~H. Shim,
``Persistent {UAV} service: An improved scheduling formulation and prototypes of system components,''
\textit{J. Intell. Robot. Syst.}, vol.~74, no.~1--2, pp.~221--232, 2014.

\bibitem{2}
P.~Maini, K.~Yu, P.~B. Sujit, and P.~Tokekar,
``Persistent monitoring with refueling on a terrain using a team of aerial and ground robots,''
in \textit{Proc. IEEE/RSJ Int. Conf. Intell. Robots Syst. (IROS)}, 2018, pp.~1--8.

\bibitem{3}
E.~Hartuv, N.~Agmon, and S.~Kraus,
``Scheduling spare drones for persistent task performance under energy constraints,''
in \textit{Proc. 17th Int. Conf. Auton. Agents Multiagent Syst. (AAMAS)}, Stockholm, Sweden, 2018, pp.~532--540.

\bibitem{4}
A.~K. Erlang,
``Solution of some problems in the theory of probabilities of significance in automatic telephone exchanges,''
\textit{Post Office Electr. Eng. J.}, vol.~10, pp.~189--197, 1917.

\bibitem{5}
E.~Galceran and M.~Carreras,
``A survey on coverage path planning for robotics,''
\textit{Robot. Auton. Syst.}, vol.~61, no.~12, pp.~1258--1276, 2013.

\bibitem{6}
G.~S. Avellar, G.~A.~S. Pereira, L.~C.~A. Pimenta, and P.~Iscold,
``Multi-{UAV} routing for area coverage and remote sensing with minimum time,''
\textit{Sensors}, vol.~15, no.~11, pp.~27783--27803, 2015.

\bibitem{7}
S.~Park, L.~Zhang, and S.~Chakraborty,
``Battery assignment and scheduling for drone delivery businesses,''
in \textit{Proc. IEEE/ACM Int. Symp. Low Power Electron. Design (ISLPED)}, 2017.

\bibitem{8}
D.~L. Jagerman,
``Some properties of the {E}rlang loss function,''
\textit{Bell Syst. Tech. J.}, vol.~53, no.~3, pp.~525--551, 1974.

\bibitem{9}
F.~P. Kelly,
\textit{Reversibility and Stochastic Networks}.
Wiley, 1979.

\bibitem{10}
W.~A. Massey and W.~Whitt,
``Stationary-process approximations for the nonstationary {E}rlang loss model,''
\textit{Oper. Res.}, vol.~44, no.~6, pp.~976--983, 1996.

\bibitem{11}
A.~Li and W.~Whitt,
``Approximate blocking probabilities in loss models with independence and distribution assumptions relaxed,''
\textit{Perform. Eval.}, vol.~80, pp.~82--101, 2014.

\bibitem{12}
C.~Bandi, D.~Bertsimas, and N.~Youssef,
``Robust queueing theory,''
\textit{Oper. Res.}, vol.~63, no.~3, pp.~676--700, 2015.

\bibitem{13}
M.~Erdelj, O.~Saif, E.~Natalizio, and I.~Fantoni,
``{UAVs} that fly forever: Uninterrupted structural inspection through automatic {UAV} replacement,''
\textit{Ad Hoc Netw.}, vol.~94, p.~101612, 2019.

\bibitem{14}
C.~Liu, J.~Zhao, and N.~Sun,
``A review of collaborative air--ground robots research,''
\textit{J. Intell. Robot. Syst.}, vol.~106, no.~60, pp.~1--28, 2022.

\bibitem{15}
G.~Shi, N.~Karapetyan, A.~B. Asghar, J.-P. Reddinger, J.~Dotterweich, J.~Humann, and P.~Tokekar,
``Risk-aware {UAV-UGV} rendezvous with chance-constrained {M}arkov decision process,''
in \textit{Proc. 61st IEEE Conf. Decis. Control (CDC)}, 2022.

\bibitem{16}
E.~B. Wilson,
``Probable inference, the law of succession, and statistical inference,''
\textit{J. Amer. Statist. Assoc.}, vol.~22, no.~158, pp.~209--212, 1927.

\bibitem{17}
A.~Agresti and B.~A. Coull,
``Approximate is better than `exact' for interval estimation of binomial proportions,''
\textit{Amer. Statist.}, vol.~52, no.~2, pp.~119--126, 1998.

\bibitem{18}
A.~Ben-Tal, L.~El~Ghaoui, and A.~Nemirovski,
\textit{Robust Optimization}.
Princeton Univ. Press, 2009.

\bibitem{19}
D.~Bertsimas and M.~Sim,
``The price of robustness,''
\textit{Oper. Res.}, vol.~52, no.~1, pp.~35--53, 2004.

\bibitem{20}
A.~Nilim and L.~El~Ghaoui,
``Robust control of {M}arkov decision processes with uncertain transition matrices,''
\textit{Oper. Res.}, vol.~53, no.~5, pp.~780--798, 2005.

\bibitem{21}
A.~B. Asghar, S.~Sundaram, and S.~L. Smith,
``Multirobot persistent monitoring: Minimizing latency and number of robots with recharging constraints,''
\textit{IEEE Trans. Robot.}, vol.~41, pp.~236--252, 2025.

\end{thebibliography}
\end{document}